\documentclass[a4paper,conference]{IEEEtran}

\usepackage{xurl}
\usepackage{cite}
\usepackage{amsmath,amssymb,amsfonts}
\usepackage{algorithmic}
\usepackage{graphicx}
\usepackage{textcomp}
\usepackage{xcolor}
\usepackage{blindtext}
\usepackage{multicol}
\usepackage{multirow}
\usepackage{array}
\usepackage{booktabs}
\usepackage{mathtools}

\usepackage[utf8]{inputenc}
\usepackage{lipsum}
\usepackage{framed}
\usepackage{color}
\usepackage{capt-of}
\usepackage{graphics}

\usepackage{colortbl} %newwwwwwww
\usepackage{url}
\usepackage{hyperref}
\usepackage{siunitx} 
\usepackage[caption=false]{subfig}

\usepackage{pifont}% http://ctan.org/pkg/pifont

\definecolor{mygray}{gray}{0.3}
\usepackage{pifont}% http://ctan.org/pkg/pifont
\newcommand*\colourcheck[1]{%
  \expandafter\newcommand\csname #1check\endcsname{\textcolor{#1}{\ding{52}}}%
}
\colourcheck{green}

\newcommand*\colourcross[1]{%
  \expandafter\newcommand\csname #1cross\endcsname{\textcolor{#1}{\ding{55}}}%
}

\newcommand*\colourstar[1]{%
  \expandafter\newcommand\csname #1star\endcsname{\textcolor{#1}{\ding{72}}}%d
}

\colourcross{blue}
\colourcross{red}
\colourstar{blue}

{ \vspace{5mm}}

\ifCLASSINFOpdf
\else
\fi

\begin{document}

\title{Zero-Shot Sketch Based Image Retrieval using Graph Transformer
}

\author{\IEEEauthorblockN{Sumrit Gupta}
\IEEEauthorblockA{\textit{Indian Institute Of Technology Bombay}\\
Mumbai, India \\
sumritg267@gmail.com}
\and
\IEEEauthorblockN{Ushasi Chaudhuri}
\IEEEauthorblockA{\textit{Indian Institute Of Technology Bombay}\\
Mumbai, India \\
ushasi2cool@gmail.com}
\and
\IEEEauthorblockN{Biplab Banerjee}
\IEEEauthorblockA{\textit{Indian Institute Of Technology Bombay}\\
Mumbai, India \\
getbiplab@gmail.com}
}

\maketitle

\begin{abstract}
The performance of a zero-shot sketch-based image retrieval (ZS-SBIR) task is primarily affected by two challenges. The substantial domain gap between image and sketch features needs to be bridged, while at the same time the side information has to be chosen tactfully. Existing literature has shown that varying the semantic side information greatly affects the performance of ZS-SBIR. To this end, we propose a novel graph transformer based zero-shot sketch-based image retrieval (GTZSR) framework for solving ZS-SBIR tasks which uses a novel graph transformer to preserve the topology of the classes in the semantic space and propagates the context-graph of the classes within the embedding features of the visual space. To bridge the domain gap between the visual features, we propose minimizing the Wasserstein distance between images and sketches in a learned domain-shared space. We also propose a novel \textit{compatibility loss} that further aligns the two visual domains by bridging the domain gap of one class with respect to the domain gap of all other classes in the training set. Experimental results obtained on the extended Sketchy, TU-Berlin, and QuickDraw datasets exhibit sharp improvements over the existing state-of-the-art methods in both ZS-SBIR and generalized ZS-SBIR. 
\end{abstract}


\IEEEpeerreviewmaketitle

\section{Introduction}
Image retrieval methods are widely used and tasks like face sketch-photo synthesis play critical role in many applications such as the digital entertainment and law enforcement. Sketch Based Image Retrieval systems are those where we use sketch as input for retrieving the images. A lot of fields such as search engines, digital library, medical diagnosis, geographical information, sensing remote systems, photo sharing sites and crime prevention are the examples where sketch based image retrieval techniques can be effective. We use graph neural networks to be able to better utilise the side information and consider both the visual information and semantic information simultaneously to narrow the domain gap between sketch and image modalities in the zero shot setting when the categories of query sketches have never been seen during training. 

%Problem of (G)ZS-SBIR.
In traditional sketch-based image retrieval (SBIR) \cite{eitz2010sketch} task, a model is trained on a number of classes and at the inference stage, the model is capable of retrieving only those trained class images. However in practical applications, it is quite possible that the trained model may encounter unseen class query samples during inference for many retrieval applications. A zero-shot learning (ZSL) \cite{xian2017zero, xian2018feature, romera2015embarrassingly,zhang2015zero,zhang2016zero,kodirov2017semantic,yang2016revisiting} experimental protocol helps to bridge this challenge by exploiting a semantic information to aid the training process and better align the seen and the unseen classes of sketch and images. Further, if the test set comprises of both the seen as well as unseen class images, we call it a generalized ZSL (G-ZSL) framework. While there has been many successful attempts to solve for the SBIR task, integration of ZSL with SBIR task (ZS-SBIR) is still a problem which needs improved solutions. In this work, we aim to solve (G)ZS-SBIR task with improved performance by looking into some of the major short-comings in the existing models.

%Challenges in (g)zs-sbir
A key challenge in solving ZS-SBIR is to align the sketch, image, and semantic text modalities in contrast to only the image and semantic text modalities usually considered for alignment in ZSL. While the sketch and images are two completely different representations of the same underlying object, there is a substantial within-category variance in the representation of each object in the sketch domain as they are drawn by various amateur artists. It has been observed from the previously proposed models such as~\cite{dutta2019semantically,dey2019doodle} that the final performance of the proposed model relies greatly on the chosen semantic space.  Another important challenge is that given the non-linear nature of the mapping between visual and semantic space, the latent space may diverge from the original semantic space thus limiting the GZS-SBIR performance. It is desirable to ensure that the latent space shares equivalent characteristics as of the original semantic space.  

A major research gap in the existing models that solves for ZS-SBIR tasks is that there is no common evaluative measure for all the papers. Some papers report their performances on precision@100 values, while some on precision@200 values, whichever gives better set of results. This makes the results incomparable with each other. 
In order to remove the inconsistencies, we prefer to put a complete set of results that should help in better comparison with all the state-of-the-art models.
%To demonstrate the efficacy of a model, it is essential to show a fair comparison with all the evaluative measure and perform equally well in all of them. 
Further, it is crucial to see how the proposed ZS-SBIR model performs on real-world application scenario by considering amateur artists sketches from all around the world. To this end, besides reporting the performance on the widely used and photo-realistic TU-Berlin and Sketchy datasets, we feel it is necessary to note the performance of the model on the large-scale QuickDraw dataset as well.
%
%The existing work using generative framework
The ZS-SBIR literature primarily tries to solve this as a domain adaptation problem. It comprises of both generative and discriminative deep-learning based models. The generative models such as  \cite{shen2018zero,yelamarthi2018zero,p2020stacked,dutta2019semantically,dutta2019style,dutta2020styleguide} tries to align the visual sketch and the photo images onto a latent semantic space using adversarial training. In ZSIH~\cite{shen2018zero}, the authors propose a generative hashing mechanism to reconstruct the semantic embeddings of each class names. The SEM-PCYC also adopted a similar approach while exploiting a cyclic-GAN framework. Style transfer~\cite{dutta2019style} seeks to bridge the domain gap between sketch and photo images to solve the problem of ZS-SBIR. It utilizes the sketch query image to generate fake image samples using image specific styles.

%The existing work using discriminative framework
On the other hand, some of the notable discriminative models involve \cite{dey2019doodle,liu2017deep,duttaadaptive,sketret}, to name a few. The Doodle2search~\cite{dey2019doodle} uses a triplet architecture and uses gradient reversal layers to enforce learning domain agnostic features from image and sketches. On the other hand, \cite{duttaadaptive} pointed out that the fall in the performance attributed to the imbalance in the class-wise samples can be tackled by using a novel adaptive margin regularizer. While the above-mentioned frameworks try to improve on the accuracy of the model by synthesizing a real-valued feature space, models such as ~\cite{dutta2019semantically,liu2017deep,liu2019semantic,chaudhuri2020simplified} propose a hashed feature space which makes the retrieval a highly efficient process. While most of the existing frameworks demonstrate their results on ZS-SBIR, only a few report their performance on the more realistic task of GZS-SBIR \cite{dutta2019semantically,dutta2020semantically,p2020stacked}.

%The existing work that use GCN
The utility of semantic space has been validated in various previously proposed frameworks, such as~\cite{dey2019doodle,dutta2019semantically,liu2019semantic,zhang2020zero}, to name a few. The visual space is aligned with the semantic space to aid the training process and learn from the topology of the semantic space. However, while~\cite{dey2019doodle,dutta2019semantically,liu2019semantic} used the semantic space, they made the semantic space latent and learnable causing the network to eventually loose the class-wise topology information as the network is trained for more number of epochs. To preserve the original topology of the semantic space, \cite{chaudhuri2020crossatnet,zhang2020zero} proposed using a graph convolution network (GCN)~\cite{kipf2016semi}. While in~\cite{chaudhuri2020crossatnet} the authors use a GCN directly on the semantic graph, in \cite{zhang2020zero} the authors create a fully-connected graph whose edge weights correspond to the semantic distances and the node features comprise of class-wise visual features. 

%what we propose: compatibility and graph transformer
In this work, to get a better semantic model we propose a graph transformer based zero-shot sketch-based image retrieval (GTZSR) framework. 
Graph transformers~\cite{yun2019graph} use multi-headed attention mechanism which takes a single instance of data and encodes its context, which in this case is the semantic graph. Hence, it captures the overall topology of each word with respect to its context graph. By using a graph transformer, we enforce the embedding feature space to retain the semantic topology information, which otherwise would have been lost by using just a conventional convolutional neural network. In addition, for an improved domain alignment, we propose a novel compatibility loss function, whose primary task is to further align the sketch and image modalities.

%what we propose: wasserstein,
Further, to bridge the domain gap between the two modalities of visual data, we propose to use the Wasserstein distance~\cite{kolouri2017optimal} to transform one distribution into another, by solving for the optimal transport problem. Wasserstein distance measures the distance between the probability distributions of sketches and photos. We used the Wasserstein distance in the visual domain because it is well-suited to this domain where an underlying similarity in the sketch and image features is more important than exactly matching the likelihoods and narrows the domain gap in between the sketch and the image features.

%Our contributions
In this paper, we solve a ZS-SBIR problem and extend it for a GZS-SBIR setup. Our primary contributions are:
\begin{itemize}
    \item We propose a novel compatibility loss measure to bridge the domain gap between the two visual modalities of image and sketches.
    \item A novel graph transformer network to preserve the context graph of the semantic space and exploit this knowledge it align the visual space.
    \item To help further align the image and sketch modalities, we propose minimizing the Wassterstein distance between the visual modalities so as to transform the distribution of images to sketches.
    \item Perform extensive experiments on the standard benchmarked datasets of the extended Sketchy~\cite{sangkloy2016sketchy}, TU-Berlin \cite{eitz2010sketch}, and QuickDraw~\cite{dey2019doodle} and outperform the existing state-of-the-art on all the evaluation metrices by $10-20\%$.
\end{itemize}
In the following section we brief about the overall proposed methodology.   

%add your block diagram!
\begin{figure*}
\centerline{\includegraphics[width=0.7\linewidth]{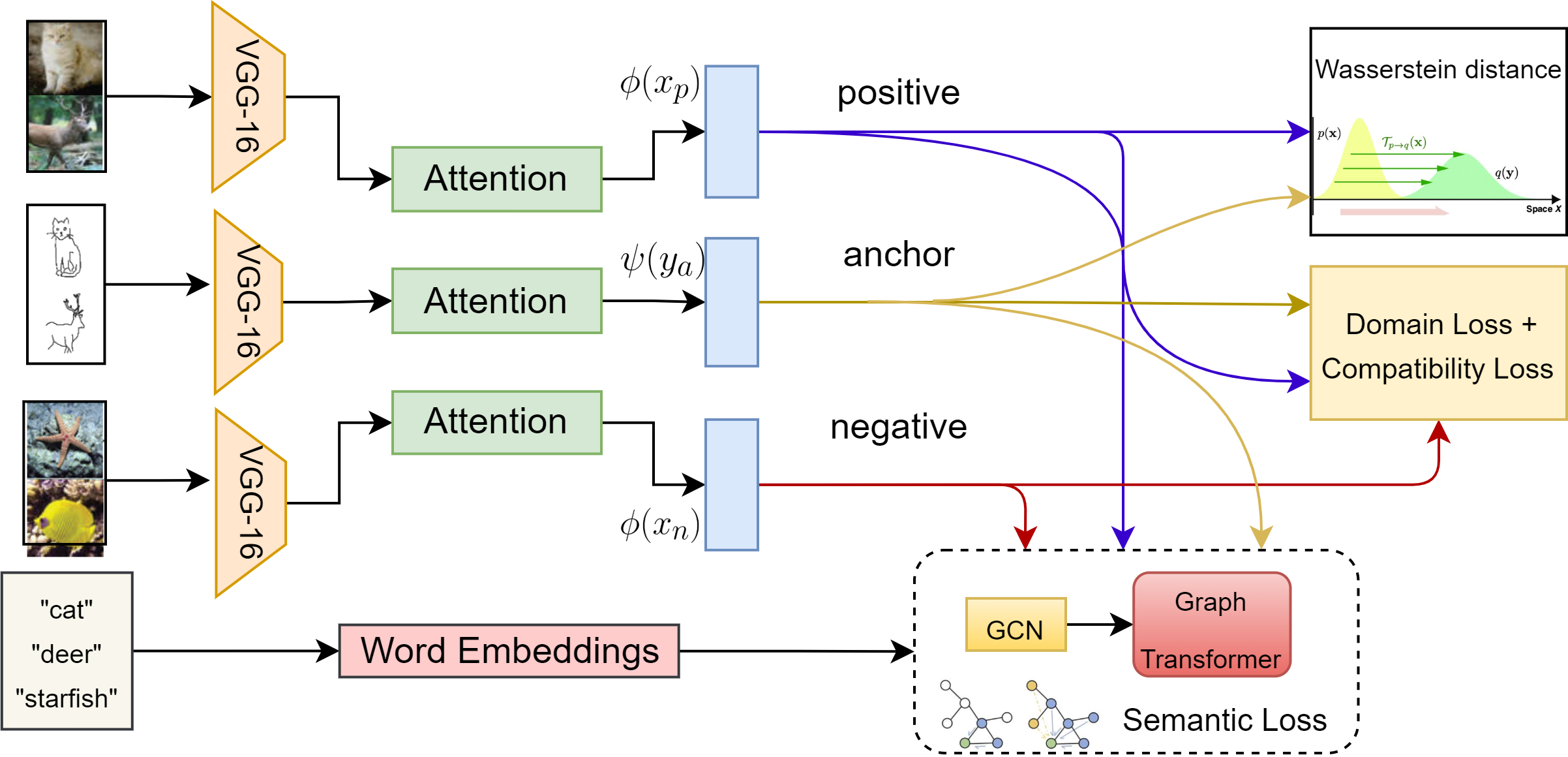}}
\caption{Overall pipeline of the proposed graph transformer based zero-shot sketch-based image retrieval (GTZSR) architecture. The encoders $\phi(\cdot)$ and $\psi(\cdot)$ encodes the images and sketches to extract meaningful features. The domain gap between the two visual space is bridged by minimizing their Wasserstein distance, domain loss, and compatibility loss. A graph transformer has been used to preserve the topology of the semantic space. }
\label{fig:blockdiag}\vspace{-2mm}
\end{figure*}

\section{Methodology}
\subsection{Problem Formulation}
%\noindent\textbf{Preliminaries.} 
Let us denote the sketches as $y_{i}$, images as $x_{i}$, and the category labels as $l_{i}$ for a dataset $P=\left\{\left(x_{i}, y_{i}, l_{i}\right) \mid l_{i} \in \mathcal{L}, x_{i} \in \mathcal{X}, y_{i} \in \mathcal{Y}\right\}$. Here, $i$ denotes the $i^{th}$ instance of the dataset with total $N$ instances. We consider the semantic embeddings of the class-names as the side information and represent it by $z_i \in \mathcal{Z}$. We then divide the dataset into a training set $D_{\text {train }}$ and a testing set $D_{t e s t}$, comprising of the seen and the unseen classes, respectively. The model is trained on the seen classes, i.e.,  data instances from $\left(l_{i} \in \mathcal{L}^{s}\right)$.  Once trained, the model is then tested on the unseen classes, sharing the labels from $\left(l_{i} \in \mathcal{L}^{u}\right)$. The seen and the unseen classes are complimentary set of each other, such that $\mathcal{L}^{s} \cap \mathcal{L}^{u}=\varnothing$. We define the set of seen and unseen photos as $\mathcal{X}^{s}=\left\{x_{i} ; l_{x}\left(x_{i}\right) \in \mathcal{L}^{s}\right\}_{i=1}^{N}$ and $\mathcal{X}^{u}=\mathcal{X} \backslash \mathcal{X}^{s}$ and the set of seen and unseen sketches as $\mathcal{Y}^{s}$ and $\mathcal{Y}^{u}$. In the ZS-SBIR setup, a sketch query is used to mine image samples from $D_{test}$, which comprises samples from $\mathcal{Y}^u$. In the 
GZS-SBIR setup, $D_{test}$ comprises samples from $\mathcal{Y}^u$ and $\mathcal{Y}^s$.

The main task for our model is to bridge the domain gap between sketches and images during the training phase, while aligning both the visual data modalities with the semantic space. This is achieved by  minimizing the Wassterstein distance and the novel compatibility loss between the visual modalities. A novel graph transformer has been used to preserve the context graph of the semantic space. The idea is to train the model on the seen classes and aligning the visual and semantic space in such a way that the model performs very well on the unseen classes at the inference phase. Fig.~\ref{fig:blockdiag} illustrates the overall pipeline of the proposed framework.% At the test stage, the model is supposed to retrieve the related images given a sketch $x$ from the testing set $D_{\text {test }}$.

\subsection{Overall GTZSR Architecture}
To realize the overall ZS-SBIR network, we primarily have two encoding networks for the visual data domains, i.e., for images and sketches. We also have a semantic encoding network that preserves the overall semantic topology of the classes. Using these feature encoders we train the network to realize the embedding feature space by minimizing an overall objective function. 

\subsubsection{Visual Encoding Networks}
The encoding framework learns two embedding functions $\phi: \mathcal{X} \rightarrow \mathbb{R}^{D}$ and $\psi: \mathcal{Y} \rightarrow \mathbb{R}^{D}$ which respectively map the photo and sketch domain into a common embedding space and later, these functions are used in the retrieval task during the test phase. We feed the model with an anchor sketch $y_a \in \mathcal{Y}$ and two photo exemplars $x_{p}, x_{n} \in \mathcal{X}$ where $x_{p}$ and $y_{a}$ belong to the same class and $x_{n}$ and $y_{a}$ belong to the different class. Here, the embeddings should fulfil the following condition: $d\left(\phi\left(x_{p}\right), \psi(y_{a})\right)<d\left(\phi\left(x_{n}\right), \psi(y_{a})\right)$, when $l_{x}\left(x_{p}\right)=l_{y}(y_{a})$ and $l_{x}\left(x_{n}\right) \neq l_{y}(y_{a})$, where $d(\cdot, \cdot)$ is a distance function which we set as $\ell_{2}$ distance. 

Since the sketches and images are from different modalities, the two embedding functions do not share the weights during training. They are guided to learn a modality-agnostic representation to narrow the domain gap. We use a soft-attention network on these visual features to learn an attention mask which assigns different weights to different regions of an image. We apply $1 \times 1$ convolution layers on the visual feature map to compute the attention mask. %For a given a feature map $f$ and an attention mask $att$, we compute the attention module's output by $f+f \cdot$ $att$,

\subsubsection{Semantic Encoding Network}
The main goal of this network is to generate meaningful representation for the side information. Here, we aim transfer the semantic topology to the visual space to be able to generalize the knowledge obtained from the seen categories to the unseen categories. Graph convolution networks (GCN)~\cite{kipf2016semi} have shown great results to transfer the knowledge between categories, hence, we use GCN to capture the semantic space. The relations among all the categories are implied by the semantic embeddings and each element $a^{i, j}$ in the adjacency matrix $A$ is decided by the semantic embeddings from $\mathcal{Z}$, which is computed by the cosine distance between semantic features and $a^{i, j}$ indicates the similarity between the node $h_{i}$ and the node $h_{j}$. The semantic relations of the nodes are obtained by the model by their similarity on the semantic space.

%Let $H^{(l)}=\left(h_{1}^{(l)}, h_{2}^{(l)}, \ldots, h_{m}^{(l)}\right)^{T}$ denotes the feature matrix of the nodes in the $l$-th GCN layer and $A \in \mathbb{R}^{m \times m}$ denotes the graph adjacency matrix. The layer-wise propagation rule in GCN is given by:
%\begin{equation}
% H^{(l+1)}=\sigma\left(\hat{A} H^{(l)} W^{(l)}\right)   
%\end{equation}

%where $\hat{A}$ is a normalized version of the graph adjacency matrix $A ; W^{(l)}$ is a parameter matrix; $\sigma$ is a non-linear operation like ReLU. 

\noindent\textbf{Graph Transformer Layer:} We add a graph transformer network (GTN) \cite{yun2019graph} layer that identifies multi-hop connections and useful meta-paths (i.e., paths connected with edges) to learn new graph structures for effectively learning node representation on graphs. GTNs try to use multiple candidate adjacency matrices to seek new graph structures and learn more powerful node representations for improved graph convolution. 

% A meta-path $p$ is a path on the graph $G$ that is connected with edges, i.e., $v_{1} \stackrel{t_{1}}{\longrightarrow} v_{2} \stackrel{t_{2}}{\longrightarrow} \ldots \stackrel{t_{l}}{\rightarrow} v_{l+1}$, where $t_{l} \in \mathcal{T}^{e}$ denotes an $l$-th edge of meta-path. Meta path defines a composite relation $R=t_{1} \circ t_{2} \ldots \circ t_{l}$ between node $v_{1}$ and $v_{l+1}$, where $x_{1} \circ x_{2}$ denotes the composition of relation $x_{1}$ and $x_{2}$. From the sequence of edge types $\left(t_{1}, t_{2}, \ldots, t_{l}\right)$, we can obtain the adjacency matrix $A_{\mathcal{P}}$ of the meta-path $P$ by the multiplications of adjacency matrices as
% $
% A_{\mathcal{P}}=A_{t_{l}} \ldots A_{t_{2}} A_{t_{1}}
% $

We define a $l$-length meta-path $P$ with vertices connected as $v_{1} \stackrel{e_{1}}{\longrightarrow} v_{2} \stackrel{e_{2}}{\longrightarrow} \ldots \stackrel{e_{l}}{\rightarrow} v_{l+1}$, and the path's $l$-th edge is denoted by $e_{l}$.    
Here a composite relation is defined between the vertices $v_{1}$ and $v_{l+1}$ as $R=e_{1} \circ e_{2} \ldots \circ e_{l}$. Once we have the composite relation, we can obtain the meta-path $P$'s adjacency matrix $A_{\mathcal{P}}$ by multiplying the adjacency matrices using the sequence of edge types $\left(e_{1}, e_{2}, \ldots, e_{l}\right)$ as
$$
A_{\mathcal{P}}=A_{e_{l}} \ldots A_{e_{2}} A_{e_{1}}
$$

\begin{figure}
\centerline{\includegraphics[width=\linewidth]{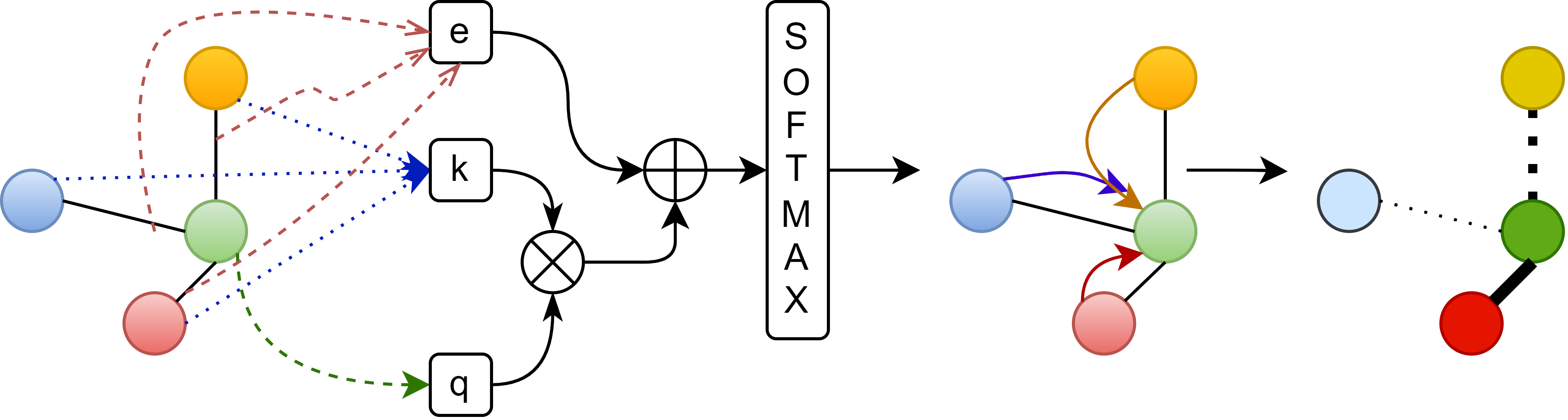}}
\vspace{-1mm}\caption{GTN performs a scaled-product of the features of query (q) of each node and its neighbours (k). The attention score for an edge is obtained by applying the softmax operation on the sum of the corresponding edge features (e) and the scaled-product and the key neighbourhood information is extracted.}
\label{fig:GTN}\vspace{-2mm}
\end{figure}

% We denote $\alpha_{t_{l}}^{(l)}$ as an attention score for edge type $t_{l}$ at \textcolor{red}{the} $l$ th GT layer obtained as illustrated in Fig 2. So, adjacency matrix generated by the $l$th GT layer, $A^{(l)}$, can be viewed as a weighted sum of all meta-paths including 1-length (original edges) to $l$-length meta-paths. The cntribution of a meta-path $t_{l}, t_{l-1}, \ldots, t_{0}$, is obtained by $\prod_{i=0}^{l} \alpha_{t_{i}}^{(i)}$.
% The weight for a meta path is an attention score and it provides the importance of the meta-path for a prediction task.

The attention score at the $l$th GT layer is obtained by the mechanism as illustrated in Fig 2. It is denoted by $\alpha_{e_{l}}^{(l)}$ for edge type $e_{l}$ and the product $\prod_{i=0}^{l} \alpha_{e_{i}}^{(i)}$ provides us the contribution of the meta-path $e_{l}, e_{l-1}, \ldots, e_{0}$ and its importance for the prediction task using the attention score as the weight of the path.
Hence, GTN learns the similarities between the sketches in the images in the semantic space using the meta path and thus adapts an attention mechanism for better prediction.

\subsection{Learning Objectives}
We used a combination of the following learning objectives for our overall loss function.
\begin{itemize}
    \item \textbf{Wasserstein Distance~\cite{kolouri2017optimal} -} It measures the distance between two probability distributions. Wasserstein metric ($ \mathcal{L}_w$) between cumulative distribution functions $\mathbb{F}$ and $\mathbb{G}$ for the sketch and photo embedding $\psi(y_a)$ and $\phi(x_p)$, respectively, is given as:
    \begin{equation}
        \mathcal{L}_w\left(\mathbb{F}, \mathbb{G}\right)=\inf _{\gamma \in \Pi\left(\mathbb{F}, \mathbb{G}\right)} \mathbb{E}_{(\psi(y_a), \phi(x_p)) \sim \gamma}[\|\psi(y_a)-\phi(x_p)\|],
    \end{equation}
    where we denote the set of all joint distributions $\gamma(\psi(y_a), \phi(x_p))$ by $\Pi\left(\mathbb{F}, \mathbb{G}\right)$ whose marginals are $\mathbb{F}$ and $\mathbb{G}$, respectively. Intuitively, wasserstein distance measures the cost of optimal transport that must be done from $\psi(y_a)$ to $\phi(x_p)$ for the transformation of the distribution $\mathbb{F}$ to the distribution $\mathbb{G}$ and the mass of this transfer is indicated by $\gamma(\psi(y_a), \phi(x_p))$. We used the wasserstein distance in the visual domain as the underlying similarity between sketch and image features are more important than exactly matching their likelihoods. Unlike other distance metrics like KL-divergence, this metric is a true probability metric and considers both the probability of and the distance between various feature spaces. Hence, it bridges the domain gap in the features of the visual space more effectively.
    
    % Here, we take infimum over all pairs of random variables (U, V) that have the cumulative distributions $\mathrm{F}$ and G, respectively. %It can also be written as
   %\begin{equation}
    %d_{p}(F, G)=\left(\int_{0}^{1}\left|F^{-1}(u)-G^{-1}(u)\right|^{p} d u\right)^{1 / p}   
   %\end{equation}

    \item \textbf{Compatibility Loss -} We propose a novel compatibility loss. It further aligns the two visual domains by bridging their domain gap between the feature embeddings of one class (from one modality) with respect to the feature embeddings of all other classes (from the other modality) in the training set. It is calculated by computing the exponential of the euclidean distance between the feature embedding of an anchor sketch and a positive class image and normalising it by the sum of this distance metric for all the classes. It is given as, 
\begin{equation}
    \mathcal{L}_{comp} = \frac{1}{N} \sum_{i=1}^{N} \frac{e^{(d(\psi\left(y_{a}\right), \phi\left(x_{p}\right)))}}{e^{(d(\psi\left(y_{a}\right), \phi\left(x_{p}\right)))} +  e^{(d(\psi\left(y_{a}\right), \phi\left(x_{n}\right)))}}
\end{equation}
%\item \textbf{Triplet Margin Loss -} 
%Further, we introduce a cross-modality triplet loss for the triads $\{\psi(y_a), \phi(x_p), \phi(x_n)\}$ to ensure that the images and sketches form class-wise dense clusters in the latent space. $\mathcal{L}_\text{cls}$ alone may not be sufficient to guarantee this, considering the high intra-class variance for both the modalities and a high degree of data imbalance inherent to the task itself. Maintaining a margin among the class boundaries helps in combating both the problems. By definition, the cross-modal triplet loss aims to bring the same class sample $\phi(x_p)$ from the image modality closer to a given sketch anchor $\psi(y_a)$ while pushing the negative image sample $\phi(x_n)$ far from $\psi(y_a)$ at least by a margin of $m$ by using the Euclidean distance metric $d$.
 %   \begin{equation}
  %   \resizebox{0.9\hsize}{!}{
  %   $\mathcal{L}_{trp}=\max \left\{d\left(\psi\left(y_{a}\right), \phi\left(x_{p}\right)\right)-d\left(\psi\left(y_{a}\right), \phi\left(x_{n}\right)\right)+m, 0\right\}$}
  %  \end{equation}
   % where, $d\left(x_{i}, y_{i}\right)=\left\|\mathbf{x}_{i}-\mathbf{y}_{i}\right\|_{p} $
    
    \item \textbf{Domain Loss -} To enforce the mapping of sketch and image samples to a common space we obtain a domain-agnostic embedding using domain loss. We make use of a fully-connected layer as a binary domain classifier $l(\cdot)$. For a given embeddings $\phi(\cdot)$ and $\psi(\cdot)$ for image and sketch, respectively,  we want to predict their domain. We set a hard decision boundaries for the domain classifier $(t$ and $1-t)$ in the min-max optimisation, where $t \in (0, 1)$.
%Following the notation, $g: \mathbb{R}^{D} \rightarrow[0,1]$ be the MLP and $b \in \mathbb{R}^{D}$ an embedding coming from the encoders network. Then let $l_{v}(b)=v \log \left(g\left(b\right)\right)+(1-v) \log \left(1-g\left(b\right)\right)$ denote the binary cross entropy of the samples, where $b$ is the embedding obtained by the encoder network and $v$ is 0 and 1 for sketch and photo domains respectively. 
Hence, the domain loss is defined as:
\begin{equation}
\resizebox{0.9\hsize}{!}{
   $\mathcal{L}_{dom}=\frac{1}{3 N} \sum_{i=1}^{N}\left((1-t)\left(\psi\left(y_{a}\right)\right)+t\left(\phi\left(x_{p}\right)\right)+t\left(\phi\left(x_{n}\right)\right)\right)$ }
\end{equation}
    
    \item \textbf{Classification Loss -} In addition to bridging the domain gaps, we also want to make the feature embeddings class-wise discernible. The standard cross-entropy loss is used for this as the classification loss and a linear classifier is connected to the shared embedding features. This ensures that the output features preserve the discriminate characters within each training category. Let $l_i$ represent the ground truth labels. This is given by,
    \begin{equation}
    \resizebox{0.9\hsize}{!}{
       $\mathcal{L}_{c l s}=-\frac{1}{N} \sum_{i=1}^{N}-\log P\left(l_{i} \mid \psi(y_a)\right) -\log P\left(l_{i} \mid \phi(x_{p/n})\right)$  }
    \end{equation}

    \item \textbf{Semantic Loss -} This loss forces the encoding of the semantic information in the obtained embedding by making use of a decoder network that reconstructs the semantic information of the corresponding category from the generated embedding by minimizing the cosine distance with the semantic representation of the category and the reconstructed feature vector. Let $c \in \mathcal{Y}^{s}$ be the corresponding category of the anchor $y_{a}$. We obtained the semantics of this category by the word2vec \cite{w2vmikolov2013distributed} embedding trained on part of Google News dataset ( $\sim 100$ billion words), ~GloVe\cite{pennington2014glove} and fastText~\cite{bojanowski2017enriching}. Let $h: \mathbb{R}^{D} \rightarrow \mathbb{R}^{300}$ be the semantic reconstruction network and $s=$ embedding $(c) \in \mathbb{R}^{300}$ be the semantics of the given category. We denote the cosine loss as $l_{c}()$ %We define the cosine loss as $l_{c}(e, s)=\frac{1}{2}\left(1-\frac{h(e) s^{v}}{\|h(e)\| \cdot\|s\|}\right)$ for a given image embedding $e \in \mathbb{R}^{D}$ 
    and the semantic loss is defined as:
   % $$
  %  \begin{aligned}
   \begin{equation}\resizebox{0.93\hsize}{!}{
   \hspace{-10mm}  $\mathcal{L}_{sem}=\frac{1}{3N}\sum_{i=1}^{N}\left(l_{c}\left(\psi\left(x_{a}\right), s_{i}\right)+l_{c}\left(\phi\left(x_{p}\right), s_{i}\right) +l_{c}\left(\phi\left(x_{n}\right), s_{i}\right)\right)$}
    \end{equation}
 %   \end{aligned}
 %   $$
\end{itemize}
The overall loss function is the sum of all the above mentioned learning objectives is given by $\mathcal{L} =   \lambda_1 \mathcal{L}_{comp} + \lambda_2\mathcal{L}_{dom} + \lambda_3\mathcal{L}_{cls} + \lambda_4\mathcal{L}_{sem} + \lambda_5\mathcal{L}_w$
%
% \begin{equation}
% \mathcal{L} =   \lambda_1 \mathcal{L}_{comp} + \lambda_2\mathcal{L}_{dom} + \lambda_3\mathcal{L}_{cls} + \lambda_4\mathcal{L}_{sem} + \lambda_5\mathcal{L}_w
% \end{equation}
Here, $\lambda_1$, $\lambda_2$, $\lambda_3$,  $\lambda_4$, and $\lambda_5$ are the hyper-parameters of the network. We use an Adam optimizer to train the network using the mini-batch stochastic gradient descent approach.

{
\setlength{\tabcolsep}{2pt}
\renewcommand{\arraystretch}{0.9}
\begin{table*}[t]\caption{{A performance comparison of our GTZSR model with SOTA on ZS-SBIR (top) and GZS-SBIR (bottom) on the extended Sketchy, TU Berlin, and Quickdraw datasets. The '-' represents the evaluation metrices which were not reported in the published papers.}} \vspace{-2mm}
\label{tab:comparison}
\scalebox{1}{
\centering\resizebox{\linewidth}{!}{
\begin{tabular}{cl|c{c}cc|cccc|cccc}
    %\hline
    %\multirow{1}{*}{} &
     &&
    % \multicolumn{2}{c}{\textbf{Sketchy-ext}  (S2)}&
    \multicolumn{4}{c}{\textbf{Sketchy-ext} }&
    \multicolumn{4}{c}{\textbf{TU Berlin-ext} } & 
    \multicolumn{4}{c}{\textbf{Quickdraw-ext} }\\
    %\multicolumn{1}{c}{\textbf{Size}}\\
    %\cline{2-5}  \cline{6-9} \cline{10-13} 
    &\textbf{Model} & \textbf{mAP}& \textbf{P@100}& \textbf{P@200}& \textbf{mAP@200} &
    \textbf{mAP}& \textbf{P@100}& \textbf{P@200}& \textbf{mAP@200}&
    \textbf{mAP}& \textbf{P@100}& \textbf{P@200}& \textbf{mAP@200}\\
    \toprule
\multirow{6}{1em}{\rotatebox{90}{\footnotesize{SBIR} }}
%&Softmax Baseline &0.114&0.172 &-&- &0.089&0.143 &-&- &-&- &-&-\\
&Siamese CNN\cite{qi2016sketch} &13.2&17.5 &-&- &10.9&14.1 &-&- &-&- &-&-\\
&SaN \cite{yu2017sketch}&11.5&12.5 &-&- &8.9&10.8 &-&- &-&- &-&-\\
&GN Triplet  \cite{sangkloy2016sketchy}&20.4&29.6 &-&- &17.5&25.3 &-&- &-&- &-&-\\
&3D shape \cite{wang2015sketch}&6.7&7.8 &-&- &5.4&6.7 &-&- &-&- &-&-\\
&DSH \cite{liu2017deep}&17.1&23.1 &-&- &12.9&28.9 &-&- &-&- &-&-\\
&GDH \cite{zhang2018generative}&18.7&25.9 &-&- &13.5&21.2 &-&- &-&- &-&-\\ \midrule
%\multirow{4}{1em}{\rotatebox{90}{\footnotesize{ZSL} }}
%&CMT  \cite{??}&0.087&0.102 &-&- &0.062&0.078 &-&- &-&- &-&-\\
%&DeViSE  \cite{??}&0.067&0.077 &-&- &0.059&0.071 &-&- &-&- &-&-\\
%&SSE \cite{zhang2015zero}&0.116&0.161 &-&- &0.089&0.121 &-&- &-&- &-&-\\
%&JLSE \cite{zhang2016zero} &0.131&0.185 &-&- &0.109&0.155 &-&- &-&- &-&-\\
%&SAE \cite{kodirov2017semantic}&0.216&0.293 &-&- &0.167&0.221 &-&- &-&- &-&-\\
%&FRWGAN \cite{??} &0.127&0.169 &-&- &0.110&0.157 &-&- &-&- &-&-\\
%&ZSH \cite{yang2016revisiting}&0.159&0.214 &-&- &0.141&0.177 &-&- &-&- &-&-\\ \hline
\multirow{15}{1em}{\rotatebox{90}{\footnotesize{ZS-SBIR}}}
&ZSIH \cite{shen2018zero} & 25.4 & 34.0 & - & - & 22.0 & 29.1 & - & - & - & -& - & - \\
%&CVAE \cite{yelamarthi2018zero} & 19.6 & - & 33.3 & 22.5 & 0.5 & - & 0.3 & 0.9 & 0.3 & - & 0.3 & 0.6 \\
&ZS-SBIR \cite{verma2019generative} & 28.9 & 35.8 & - & - & 23.8 & 33.4 & - & - & - & -& - & - \\
&SEM-PCYC \cite{dutta2019semantically} & 34.9 & 46.3 & - & - & 29.7 & 42.6 & - & - & - & -& - & - \\
&Doodle2Search \cite{dey2019doodle} & 36.9 & - & 37.0 & 46.1 & 10.9 & - & 12.1 & 15.7 & 7.5 & - & 6.7 & 9.1 \\
&SkechGCN ~\cite{zhang2020zero}& 38.2 & 53.8 &  48.7 & 56.8 & 32.4 & 50.5 & 47.8 & 52.8 & - & -& - & - \\
&StyleGuide~\cite{dutta2020styleguide} &37.5&48.4 &-&- &25.4&35.5 &-&- &-&- &-&-\\
&SAKE~\cite{liu2019semantic} &- &- &{59.8} &49.7 &47.5&59.9 &-&- &-&- &-&-\\
&STRAD (Double) ~\cite{li2019zero} &- &- &50.2 &37.9 &- &- &24.5 &15.4 &-&- &12.6 &5.4\\
&SBTKNet~\cite{tursun2021efficient} &- &- &59.6 &50.2 &48.0 &60.8 &-&- &-&- &-&-\\
&SEM-PCYC+AMDReg \cite{duttaadaptive}&39.7 &49.4 &-&- &33.0 &47.3 &-&- &-&- &-&-\\
&StyleGuide+AMDReg \cite{duttaadaptive}&41.0 &51.2 &-&- &29.1 &37.6 &-&- &-&- &-&-\\
&SAKE+AMDReg \cite{duttaadaptive}&{55.1} &{71.5} &-&- &44.7 &57.4 &-&- &-&- &-&-\\
&PCSN \cite{deng2020progressive}&52.3 &61.6 &-&- &42.4 & 51.7 &-&- &-&-  &-&-\\
&OCEAN ~\cite{zhu2020ocean} &- &- &54.9 &{57.9} &- &- &39.8 &42.7 &-&- &- &-\\
&BDA-SketRet~\cite{sketret} & 43.5 &51.2 &45.8 &55.6&  37.4& 50.4& 43.8& 54.4&  15.4 &44.0& 35.5 &34.6 \\
&\textbf{GTZSR (Ours)} &  \textbf{68.6} & \textbf{72.5} & \textbf{72.6} & 
\textbf{69.4} & \textbf{70.3} & \textbf{77.9} & \textbf{79.0} & \textbf{77.7} & \textbf{24.1} & \textbf{45.6} & \textbf{40.1} & \textbf{51.8} \\
\midrule
\multirow{9}{1em}{\rotatebox{90}{\footnotesize{GZS-SBIR}}}
&ZSIH~\cite{shen2018zero}&21.9&29.6 &-&- &14.2&21.8 &-&- &-&- &-&-\\
&ZS-SBIR~\cite{yelamarthi2018zero}&14.6&19.0 &-&- &30.1&10.2 &-&- &20.3&10.5 &-&-\\
&SEM-PCYC \cite{dutta2019semantically}&30.7&36.4 &-&-&19.2&29.8 &-&-&14.0&22.1 &-&-\\
&SEM-PCYC+AMDReg \cite{duttaadaptive}&32.0&39.8 &-&- &24.5&30.3 &-&- &-&- &-&-\\
&Style-guide \cite{dutta2019style}&33.0&38.1 &-&-&14.9& 22.6&-&- &-&-  &-&-\\
%&StyleGuide + Openmax ND ~\cite{dutta2020styleguide} &33.8&37.7 &-&-&15.1& 22.9&-&- &-&-  &-&-\\
&StyleGuide + AE ND ~\cite{dutta2020styleguide} &35.0&40.3 &-&-&15.3& 23.5&-&- &-&-  &-&-\\
&OCEAN ~\cite{zhu2020ocean} &- &- &44.3 &54.7 &- &- &31.9 &36.9 &-&- &- &-\\
&BDA-SketRet~\cite{sketret} & 22.7 &25.1 &22.6 &33.7& 25.1& 35.7 &32.9& 33.3& 15.4 &28.6 &29.5 &27.4\\
&\textbf{GTZSR (Ours)}& \textbf{61.7} & \textbf{64.0} & \textbf{65.1} & 
\textbf{62.5} & \textbf{62.8} & \textbf{66.8} & \textbf{68.0} & \textbf{67.4} & \textbf{20.4} & \textbf{39.6} & \textbf{35.5 } & \textbf{42.1} \\\bottomrule
\end{tabular}%\addtolength{\tabcolsep}{1pt}}
}
}
\end{table*}
}

\section{Experiments}
%\subsection{Dataset and Implementation Details}
\noindent\textbf{Datasets.} We use the standard benchmarked Sketchy-extended~\cite{sangkloy2016sketchy}, TU-Berlin-extended~\cite{eitz2010sketch} and Quickdraw-extended~\cite{dey2019doodle} datasets for our experiments. 
Sketchy consists of 125 categories out of which we used the 104:21 train:test split as mentioned in~\cite{yelamarthi2018zero}, which ensures that the testing categories do not previously appear in the
ImageNet dataset during the pretraining stage. %There are 100 images and at least 600 sketches in each category and this dataset was extended by researchers before by collecting 60,502 natural images from ImageNet (Deng et al. 2009) and thus, Sketchy-Extended totally contains 73,002 images and 75,479 sketches. We had split the dataset
%into 104 categories for training and 21 categories for testing and we have ensured that the testing categories do not appear in the 1,000 categories of ImageNet.
The TU-Berlin-Extended~\cite{eitz2010sketch} consists of 250 categories of images and sketches out of which we randomly choose 30 classes for testing.
%, totally contains 20,000 unique sketches and they are evenly distributed over 250 categories. We used extended version of this dataset which wasa extended by Liu et al. (2017) and they collected 204,489 images. We randomly choose 30 categories that contain at least 400 images for testing and the rest 220 categories for training. 
The Quickdraw-extended dataset~\cite{dey2019doodle} consists of 330,000 sketches and 204,000 photos in total spanning across 110 categories, where we used 80 categories for training and 30 for testing.

\noindent\textbf{Implementation Details.} We created the adjacency matrix for the semantic features for the Sketchy, TUB, and Quickdraw dataset by using the cosine distance as the weight of the edges. 
We used VGG-16~\cite{simonyan2014very} pre-trained to model the encoding networks.
%We used VGG16 as an encoding network for sketch and image, mapping the sketches and images into an embedding feature space. 
This encoding network tries to embed the images and the sketches into a common semantic space. Then we use the semantic preserving network, which takes the features embedded using the encoding network as input and utilizes the side information to force the sketch and images to maintain their category level relations. The 300-D word embeddings were created using word2vec for each class name and were used as the side information. Side information is essential since we have a ZSL task, where not all the classes are present in the training set. We build a graph convolutional network to effectively preserve the semantic topology of the classes. We used three fully-connected layers with dimensions 256, followed by ReLU activation.  To train the model, we used a batch size of 4, learning rate of $\mathbf{10^{-4}}$, momentum value of 0.9, and a decay value of 0.0005 and trained it for 75 epochs. We assigned a weight ($\lambda$ values) of 0.25 each to the semantic loss, classification loss, domain loss and compatibility loss by tuning the hyper-parameters using grid search.% and cross-validation. %A margin value of 2 was used for the triplet loss. %  \textcolor{red}{how many epochs?, margin values? 

% \begin{table}[htbp]
% \caption{Table Type Styles}
% \begin{center}
% \begin{tabular}{|c|c|c|c|}
% \hline
% \textbf{Table}&\multicolumn{3}{|c|}{\textbf{Table Column Head}} \\
% \cline{2-4} 
% \textbf{Head} & \textbf{\textit{Table column subhead}}& \textbf{\textit{Subhead}}& \textbf{\textit{Subhead}} \\
% \hline
% copy& More table copy$^{\mathrm{a}}$& &  \\
% \hline
% \multicolumn{4}{l}{$^{\mathrm{a}}$Sample of a Table footnote.}
% \end{tabular}
% \label{tab1}
% \end{center}
% \end{table}

\begin{figure*}
%\begin{figure}\center
\addtolength{\tabcolsep}{-6.5pt} 
 \scalebox{1.08}{\begin{tabular}{ccccccccc c ccccccccc}
   {\includegraphics[width=9mm, height = 9mm]{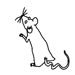}}&
   {\includegraphics[width=9mm, height = 9mm]{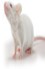}}&
   {\includegraphics[width=9mm, height = 9mm]{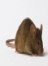}}&
   {\includegraphics[width=9mm, height = 9mm]{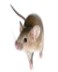}}&
   {\includegraphics[width=9mm, height = 9mm]{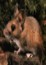}}&
   {\includegraphics[width=9mm, height = 9mm]{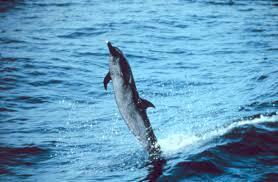}}&
   {\includegraphics[width=9mm, height = 9mm]{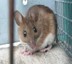}}&
   {\includegraphics[width=9mm, height = 9mm]{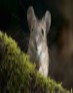}}&
   {\includegraphics[width=9mm, height = 9mm]{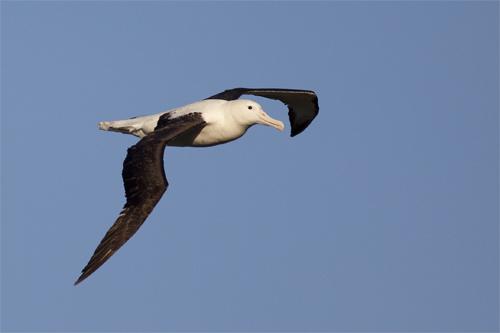}}&
\hspace{5pt}&\hspace{5pt}
   {\includegraphics[width=9mm, height = 9mm]{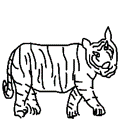}}&
   {\includegraphics[width=9mm, height = 9mm]{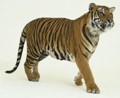}}&
   {\includegraphics[width=9mm, height = 9mm]{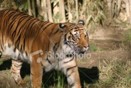}}&
   {\includegraphics[width=9mm, height = 9mm]{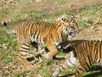}}&
   {\includegraphics[width=9mm, height = 9mm]{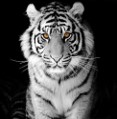}}&
   {\includegraphics[width=9mm, height = 9mm]{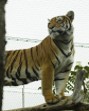}}&
   {\includegraphics[width=9mm, height = 9mm]{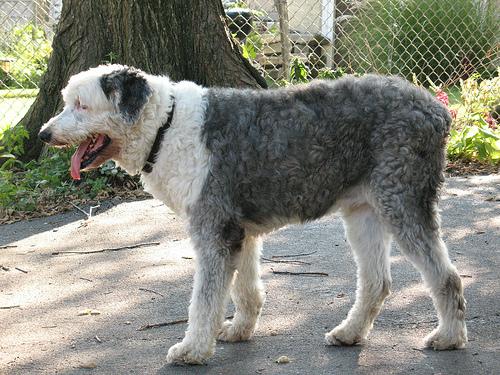}}&
   {\includegraphics[width=9mm, height = 9mm]{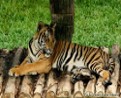}}&
   {\includegraphics[width=9mm, height = 9mm]{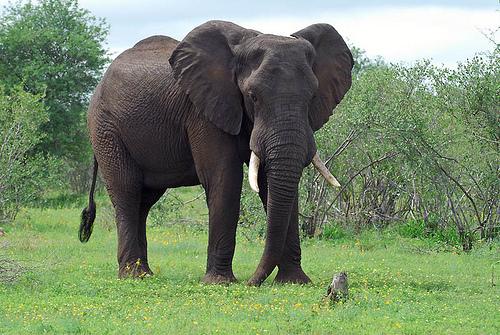}}\\
   
    \scriptsize{\tt Mouse}& \greencheck & \greencheck &\greencheck &\greencheck &\redcross &\greencheck &\greencheck &\redcross  & 
        
     &\scriptsize{\tt Tiger}& \greencheck & \greencheck &\greencheck &\greencheck &\greencheck &\redcross &\greencheck &\redcross  \\

   {\includegraphics[width=9mm, height = 9mm]{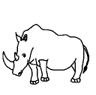}}&
   {\includegraphics[width=9mm, height = 9mm]{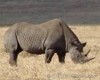}}&
   {\includegraphics[width=9mm, height = 9mm]{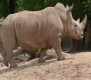}}&
   {\includegraphics[width=9mm, height = 9mm]{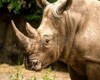}}&
   {\includegraphics[width=9mm, height = 9mm]{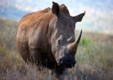}}&
   {\includegraphics[width=9mm, height = 9mm]{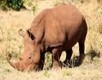}}&
   {\includegraphics[width=9mm, height = 9mm]{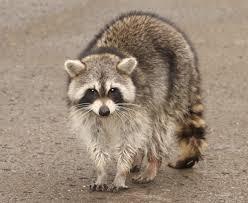}}&
   {\includegraphics[width=9mm, height = 9mm]{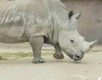}}&
   {\includegraphics[width=9mm, height = 9mm]{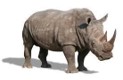}}&
   \hspace{5pt} & \hspace{5pt}
   {\includegraphics[width=9mm, height = 9mm]{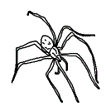}}&
   {\includegraphics[width=9mm, height = 9mm]{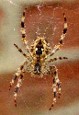}}&
   {\includegraphics[width=9mm, height = 9mm]{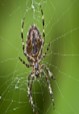}}&
   {\includegraphics[width=9mm, height = 9mm]{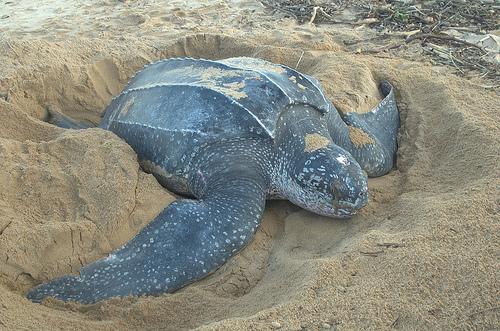}}&
   {\includegraphics[width=9mm, height = 9mm]{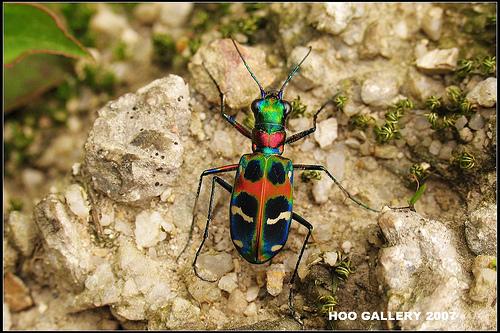}}&
   {\includegraphics[width=9mm, height = 9mm]{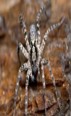}}&
   {\includegraphics[width=9mm, height = 9mm]{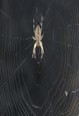}}&
   {\includegraphics[width=9mm, height = 9mm]{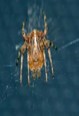}}&
   {\includegraphics[width=9mm, height = 9mm]{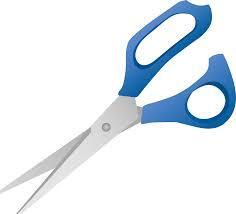}}\\  

     \scriptsize{\tt Rhinoceros}& \greencheck & \greencheck &\greencheck &\greencheck &\greencheck &\redcross &\greencheck &\greencheck  & 
     
   & \scriptsize{\tt Spider}& \greencheck & \greencheck &\redcross &\redcross &\greencheck &\greencheck &\greencheck &\redcross  \\
   
   {\includegraphics[width=9mm, height = 9mm]{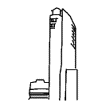}}&
   {\includegraphics[width=9mm, height = 9mm]{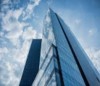}}&
   {\includegraphics[width=9mm, height = 9mm]{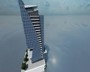}}&
   {\includegraphics[width=9mm, height = 9mm]{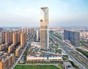}}&
   {\includegraphics[width=9mm, height = 9mm]{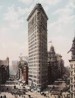}}&
   {\includegraphics[width=9mm, height = 9mm]{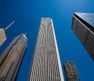}}&
   {\includegraphics[width=9mm, height = 9mm]{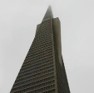}}&
   {\includegraphics[width=9mm, height = 9mm]{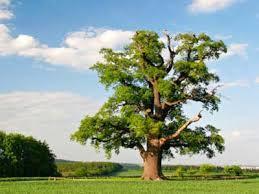}}&
   {\includegraphics[width=9mm, height = 9mm]{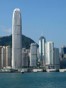}}&
     \hspace{5pt}&\hspace{5pt}
   {\includegraphics[width=9mm, height = 9mm]{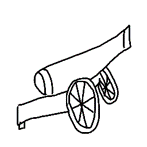}}&
   {\includegraphics[width=9mm, height = 9mm]{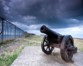}}&
   {\includegraphics[width=9mm, height = 9mm]{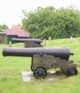}}&
   {\includegraphics[width=9mm, height = 9mm]{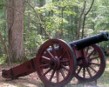}}&
   {\includegraphics[width=9mm, height = 9mm]{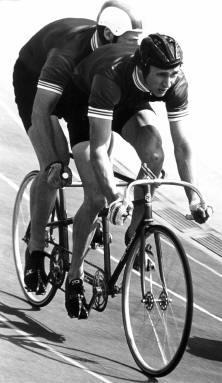}}&
   {\includegraphics[width=9mm, height = 9mm]{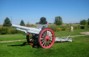}}&
   {\includegraphics[width=9mm, height = 9mm]{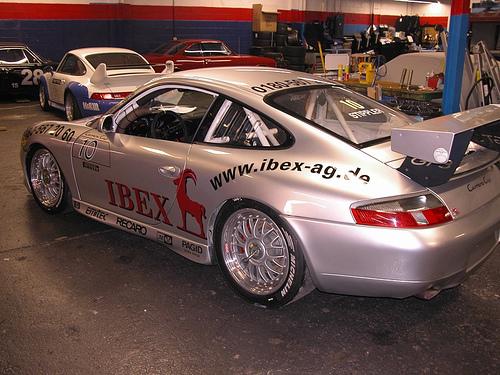}}&
   {\includegraphics[width=9mm, height = 9mm]{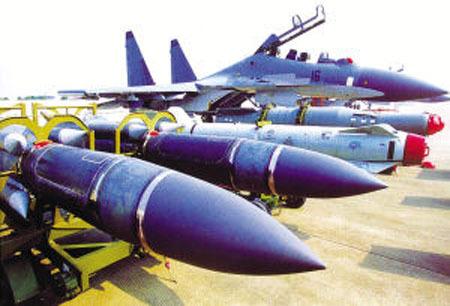}}&
   {\includegraphics[width=9mm, height = 9mm]{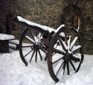}}\\

  \scriptsize{\tt Skyscrapers}& \greencheck & \greencheck &\greencheck &\greencheck &\greencheck &\greencheck &\redcross &\greencheck  & 
      
    &~\scriptsize{\tt Canon}& \greencheck & \greencheck &\greencheck &\redcross &\greencheck &\redcross &\redcross &\greencheck  \\  
  \end{tabular}}\addtolength{\tabcolsep}{6pt}  
%\end{figure}
    \caption{Top-8 retrieved images by our GTZSR-Net for some sketch queries from the Sketchy dataset for ZS-SBIR on the left and for GZS-SBIR on the right. %The correctly retrieved class images are denoted by the green check and the incorrectly retrieved class images are denoted by red crosses. 
}\vspace{-2mm}
\label{fig:hubs} 
\end{figure*}

\noindent\textbf{Training and Evaluation Protocol.} To train the network, a random sketch anchor was chosen from the training set and a corresponding class positive exemplar of image and a negative class exemplar of image were chosen as the triplets for training. The negative exemplars were hard-mined samples, whose feature embedding lied nearest to that of the positive class. To adjust for the class-imbalance of samples in the datasets, we considered similar number of anchor sketches from each training class. The model was evaluated using the mean average precision (mAP) and precision at X (P@X) values. We report the performance of the model with mAP, mAP@200, P@100, and P@200 values. \\
% Precision (P) and mean average precision (mAP) are two main metrics for evaluating the ranked retrieval results for testing queries in related SBIR studies. Precision is calculated for the top $k$ (i.e., $100,200)$ ranked results, and mAP values are calculated for the top $K$ or all ranked results. 
The $\mathrm{P}@\mathrm{K}$ is equal to the percentage of relevant images in the top $K$ ranked retrieved images. We calculate the AP values of each query using $\mathrm{P}@\mathrm{K}$ as $A P @ K=\sum_{i=1}^{K} \frac{P @ i \times \gamma(i)}{N}$. %follows:
% $$
% A P @ K=\sum_{i=1}^{K} \frac{P @ i \times \gamma(i)}{N}
% $$
where $N$ is total number of relevant images and $\gamma(i)$ is 1 if the $i$th ranked image is from the same class as the input sketch, otherwise 0. $\mathrm{mAP} @ \mathrm{K}$ is the mean $\mathrm{AP} @ \mathrm{K}$ of all queries.

\section{Results and Discussions}
We compare the performance of the proposed model with some of the existing state-of-the-art frameworks of ~\cite{sketret,shen2018zero, yelamarthi2018zero,verma2019generative,dey2019doodle,dutta2019semantically,zhang2020zero,dutta2019style,duttaadaptive}. We also lay down the performances of some of the notable works in SBIR that uses the same datasets to show how the proposed model which solves a more challenging ZS-SBIR problem achieves comparable performance. The train test split has also been made as random as possible to avoid any bias induced due to the training process. We report the ZS-SBIR and the GZS-SBIR performance of our model on all the four evaluation metrices on the three different datasets in Table~\ref{tab:comparison}.

From Table~\ref{tab:comparison} is can be seen that the proposed framework not only beats the existing state-of-the-art ZS-SBIR models, it also outperforms the existing literature in the challenging GZS-SBIR problem. Further, it shows comparable performance with the much simpler SBIR problem. Fig.~\ref{fig:hubs} illustrates the top-8 retrieved samples given a sketch query from the unseen class.  The green check marks denote correctly retrieved samples while red crosses denote incorrectly retrieved samples.

\subsection{Ablation Studies}
In this section, we carefully study the contribution of each of the loss functions and the model components and validate their significance by experimenting on the TU-Berlin dataset. Table~\ref{tab:ablation} reports the performance of various ablation studies that have been conducted.

We first perform the ablation of loss terms by removing one loss function at a time from the overall objective function. %First we train the entire network by removing $\mathcal{L}_{trp}$. A sharp fall of $\approx 13\%$ is observed in the mAP value. This validates the significance of $\mathcal{L}_{trp}$ in making the clusters of the two modalities class-wise discernible. 
First, we train the model using the full objective function minus the $\mathcal{L}_{comp}$. A further fall is observed in the overall mAP value, validating its role in bridging the domain gap between the two visual modalities. The Wasserstein distance was minimized between the image and sketch modalities to further align them well in the embedding space and to bridge their domain gap. When we remove the Wasserstein distance from the overall learning objective, a marginal fall in the performance is observed, resulting with a mAP value of 62.1. Then, we train the entire network by removing $\mathcal{L}_{cls}$. A fall of $\approx 4\%$ is observed in the mAP value. This validates the significance of $\mathcal{L}_{cls}$ in making the clusters of the two modalities class-wise discernible.
Finally, we train the model without using the $\mathcal{L}_{dom}$ that helps to align the sketch and image modalities by bridging the domain gap of their embedding feature space. A clear downfall in all the evaluation metrices validate its functionality in the proposed model.

Next, we go on to study the significance of each of the network components of the proposed model. When the model is trained without the attention networks for the images and sketches, a marginal fall in the overall performance is observed. Removal of the proposed novel graph transformer layer greatly affects the performance and results in a fall of $\approx 22\%$ in the mAP value. While on the other hand, retaining just the GCN with triplet loss as similar to the architecture of \cite{dey2019doodle} also results into a fall in the performance by a great margin. Hence, together the GCN layer along with the graph transformer on the semantic space effectively helps in preserving the semantic topology, capturing the class-wise context graph. The full model with all the learning objectives and network components outperforms the remaining network configuration. 
{
\setlength{\tabcolsep}{3pt}
\renewcommand{\arraystretch}{0.85}
\begin{table}\caption{{Ablation study for the proposed model on TU Berlin-ext.}}\vspace{-2mm}
\label{tab:ablation}
\centering\resizebox{\linewidth}{!}{
\begin{tabular}{ll{c}cccc}
    % &
    %&\multicolumn{4}{c}{\textbf{TU Berlin-ext} } & \\
    & \textbf{Model} & \textbf{mAP}& \textbf{P@100}& \textbf{P@200}& \textbf{mAP@200}\\
    %  \textbf{mAP}& \textbf{P@100}& \textbf{P@200}& \textbf{mAP@200}&
    %  \textbf{mAP}& \textbf{P@100}& \textbf{P@200}& \textbf{mAP@200}&\\
    \toprule
\multirow{4}{1em}{\rotatebox{90}{\footnotesize{Losses}}}
%&W/o $\mathcal{L}_{trp}$ & 57.1 & 67.3 & 64.2 & 68.1\\
&W/o $\mathcal{L}_{comp}$ & 54.3 & 64.4 & 62.5 & 69.4\\
&W/o Wasserstein Distance & 62.1 & 69.2 & 70.2 & 65.9\\
&W/o $\mathcal{L}_{dom}$ & 57.5	& 68.2 & 60.5 & 64.1\\
&W/o $\mathcal{L}_{cls}$ & 66.8	& 73.7 & 75.2 & 73.9\\
\midrule
\multirow{4}{1em}{\rotatebox{90}{\footnotesize{Model}}}
&W/o Attention & 67.0 & 73.3 & 75.1 & 74.2\\
&W/o Graph Transformer & 48.3 & 62.2 & 61.5 & 58.6\\
&Only GCN and $\mathcal{L}_{comp}$ & 49.3 & 58.2 & 53.9 & 59.5\\
&\textbf{Full Model} & \textbf{70.3} & \textbf{77.9} & \textbf{79.0} & \textbf{77.7} \\ \bottomrule
\end{tabular}%\addtolength{\tabcolsep}{1pt}}
}\vspace{-2mm}
\end{table}
}

\section{Conclusion \& Future Directions}
We propose a novel GTZSR framework for solving (G)ZS-SBIR tasks. It comprises of convolutional networks for encoding the visual space and graph transformer network to preserve the semantic space. We use the wasserstein distance metric to bridge the domain gap between sketch and image features and instead of reducing the domain gap between individual data instances, we bring the probability distributions of both these domains closer to each other. We also propose a novel compatibility loss which brings the embeddings of one class (from one modality) with respect to the feature embeddings of all other classes (from the other modality) in the training set. Together with wasserstein loss, we are able to bridge the domain gap problem successfully. We see that our method outperforms the existing state-of-the-art models on both the (G)ZS-SBIR tasks by quite a large margin on three large-scale benchmarked datasets. In future, the sketch based image retrieval task can also be extended to a multi-labeled setup, like the presence of a car and a dog in the same image. The task then changes to multi-class zero-shot sketch-based image retrieval. In such case, a modified cross-triplet loss~\cite{chaudhuri2021interband} or a quadruplet loss~\cite{chen2017beyond} could be found to be useful, as it can handle multi-class classification in better way by sampling more negative images for the classes.

\bibliographystyle{IEEEtran}
\bibliography{IEEEexample}

\begin{thebibliography}{10}
\providecommand{\url}[1]{#1}
\csname url@samestyle\endcsname
\providecommand{\newblock}{\relax}
\providecommand{\bibinfo}[2]{#2}
\providecommand{\BIBentrySTDinterwordspacing}{\spaceskip=0pt\relax}
\providecommand{\BIBentryALTinterwordstretchfactor}{4}
\providecommand{\BIBentryALTinterwordspacing}{\spaceskip=\fontdimen2\font plus
\BIBentryALTinterwordstretchfactor\fontdimen3\font minus
  \fontdimen4\font\relax}
\providecommand{\BIBforeignlanguage}[2]{{%
\expandafter\ifx\csname l@#1\endcsname\relax
\typeout{** WARNING: IEEEtran.bst: No hyphenation pattern has been}%
\typeout{** loaded for the language `#1'. Using the pattern for}%
\typeout{** the default language instead.}%
\else
\language=\csname l@#1\endcsname
\fi
#2}}
\providecommand{\BIBdecl}{\relax}
\BIBdecl

\bibitem{eitz2010sketch}
M.~Eitz, K.~Hildebrand, T.~Boubekeur, and M.~Alexa, ``Sketch-based image
  retrieval: Benchmark and bag-of-features descriptors,'' \emph{IEEE TVCG},
  2010.

\bibitem{xian2017zero}
Y.~Xian, B.~Schiele, and Z.~Akata, ``Zero-shot learning-the good, the bad and
  the ugly,'' in \emph{CVPR}, 2017.

\bibitem{xian2018feature}
Y.~Xian, T.~Lorenz, B.~Schiele, and Z.~Akata, ``Feature generating networks for
  zero-shot learning,'' in \emph{CVPR}, 2018.

\bibitem{romera2015embarrassingly}
B.~Romera-Paredes and P.~Torr, ``An embarrassingly simple approach to zero-shot
  learning,'' in \emph{ICML}, 2015.

\bibitem{zhang2015zero}
Z.~Zhang and V.~Saligrama, ``Zero-shot learning via semantic similarity
  embedding,'' in \emph{Proceedings of the IEEE international conference on
  computer vision}, 2015, pp. 4166--4174.

\bibitem{zhang2016zero}
------, ``Zero-shot learning via joint latent similarity embedding,'' in
  \emph{Proceedings of the IEEE Conference on Computer Vision and Pattern
  Recognition}, 2016, pp. 6034--6042.

\bibitem{kodirov2017semantic}
E.~Kodirov, T.~Xiang, and S.~Gong, ``Semantic autoencoder for zero-shot
  learning,'' in \emph{Proceedings of the IEEE Conference on Computer Vision
  and Pattern Recognition}, 2017, pp. 3174--3183.

\bibitem{yang2016revisiting}
Z.~Yang, W.~Cohen, and S.~Ruslan, ``Revisiting semi-supervised learning with
  graph embeddings,'' \emph{arXiv preprint arXiv:1603.08861}, 2016.

\bibitem{dutta2019semantically}
A.~Dutta and Z.~Akata, ``Semantically tied paired cycle consistency for
  zero-shot sketch-based image retrieval,'' in \emph{CVPR}, 2019.

\bibitem{dey2019doodle}
S.~Dey, P.~Riba, A.~Dutta, J.~Llados, and Y.-Z. Song, ``Doodle to search:
  Practical zero-shot sketch-based image retrieval,'' in \emph{CVPR}, 2019.

\bibitem{shen2018zero}
Y.~Shen, L.~Liu, F.~Shen, and L.~Shao, ``Zero-shot sketch-image hashing,'' in
  \emph{CVPR}, 2018.

\bibitem{yelamarthi2018zero}
S.~K. Yelamarthi, S.~K. Reddy, A.~Mishra, and A.~Mittal, ``A zero-shot
  framework for sketch based image retrieval,'' in \emph{ECCV}, 2018.

\bibitem{p2020stacked}
A.~Pandey, A.~Mishra, V.~K. Verma, A.~Mittal, and H.~Murthy, ``Stacked
  adversarial network for zero-shot sketch based image retrieval,'' in
  \emph{WACV}, 2020.

\bibitem{dutta2019style}
T.~Dutta and S.~Biswas, ``Style-guided zero-shot sketch-based image
  retrieval.'' in \emph{BMVC}, 2019.

\bibitem{dutta2020styleguide}
T.~{Dutta}, A.~{Singh}, and S.~{Biswas}, ``Styleguide: Zero-shot sketch-based
  image retrieval using style-guided image generation,'' \emph{IEEE TM}, 2020.

\bibitem{liu2017deep}
L.~Liu, F.~Shen, Y.~Shen, X.~Liu, and L.~Shao, ``Deep sketch hashing: Fast
  free-hand sketch-based image retrieval,'' in \emph{CVPR}, 2017.

\bibitem{duttaadaptive}
T.~Dutta, A.~Singh, and S.~Biswas, ``Adaptive margin diversity regularizer for
  handling data imbalance in zero-shot sbir,'' in \emph{ECCV}, 2020.

\bibitem{sketret}
U.~Chaudhuri, R.~Chavan, B.~Banerjee, A.~Dutta, and Z.~Akata, ``Bda-sketret:
  Bi-level domain adaptation for zero-shot sbir,'' \emph{arXiv preprint
  arXiv:2201.06570v1}, 2022.

\bibitem{liu2019semantic}
Q.~Liu, L.~Xie, H.~Wang, and A.~L. Yuille, ``Semantic-aware knowledge
  preservation for zero-shot sketch-based image retrieval,'' in \emph{ICCV},
  2019.

\bibitem{chaudhuri2020simplified}
U.~Chaudhuri, B.~Banerjee, A.~Bhattacharya, and M.~Datcu, ``A simplified
  framework for zero-shot cross-modal sketch data retrieval,'' in \emph{CVPRW},
  2020.

\bibitem{dutta2020semantically}
A.~Dutta and Z.~Akata, ``Semantically tied paired cycle consistency for
  any-shot sketch-based image retrieval,'' \emph{IJCV}, 2020.

\bibitem{zhang2020zero}
Z.~Zhang, Y.~Zhang, R.~Feng, T.~Zhang, and W.~Fan, ``Zero-shot sketch-based
  image retrieval via graph convolution network.'' in \emph{AAAI}, 2020.

\bibitem{chaudhuri2020crossatnet}
U.~Chaudhuri, B.~Banerjee, A.~Bhattacharya, and M.~Datcu, ``Crossatnet-a novel
  cross-attention based framework for sketch-based image retrieval,''
  \emph{Image and Vision Computing}, 2020.

\bibitem{kipf2016semi}
T.~N. Kipf and M.~Welling, ``Semi-supervised classification with graph
  convolutional networks,'' in \emph{ICLR}, 2016.

\bibitem{yun2019graph}
S.~Yun, M.~Jeong, R.~Kim, J.~Kang, and H.~J. Kim, ``Graph transformer
  networks,'' \emph{NeurIPS}, vol.~32, pp. 11\,983--11\,993, 2019.

\bibitem{kolouri2017optimal}
S.~Kolouri, S.~R. Park, M.~Thorpe, D.~Slepcev, and G.~K. Rohde, ``Optimal mass
  transport: Signal processing and machine-learning applications,'' \emph{IEEE
  signal processing magazine}, vol.~34, no.~4, pp. 43--59, 2017.

\bibitem{sangkloy2016sketchy}
P.~Sangkloy, N.~Burnell, C.~Ham, and J.~Hays, ``The sketchy database: learning
  to retrieve badly drawn bunnies,'' \emph{ACM TOG}, 2016.

\bibitem{w2vmikolov2013distributed}
T.~Mikolov, I.~Sutskever, K.~Chen, G.~S. Corrado, and J.~Dean, ``Distributed
  representations of words and phrases and their compositionality,'' in
  \emph{NeurIPS}, 2013, pp. 3111--3119.

\bibitem{pennington2014glove}
J.~Pennington, R.~Socher, and C.~D. Manning, ``Glove: Global vectors for word
  representation,'' in \emph{EMNLP}, 2014.

\bibitem{bojanowski2017enriching}
P.~Bojanowski, E.~Grave, A.~Joulin, and T.~Mikolov, ``Enriching word vectors
  with subword information,'' \emph{ACL}, 2017.

\bibitem{qi2016sketch}
Y.~Qi, Y.-Z. Song, H.~Zhang, and J.~Liu, ``Sketch-based image retrieval via
  siamese convolutional neural network,'' in \emph{ICIP}, 2016.

\bibitem{yu2017sketch}
Q.~Yu, Y.~Yang, F.~Liu, Y.-Z. Song, T.~Xiang, and T.~M. Hospedales,
  ``Sketch-a-net: A deep neural network that beats humans,'' \emph{IJCV}, 2017.

\bibitem{wang2015sketch}
F.~Wang, L.~Kang, and Y.~Li, ``Sketch-based 3d shape retrieval using
  convolutional neural networks,'' in \emph{Proceedings of the IEEE Conference
  on Computer Vision and Pattern Recognition}, 2015, pp. 1875--1883.

\bibitem{zhang2018generative}
J.~Zhang, F.~Shen, L.~Liu, F.~Zhu, M.~Yu, L.~Shao, H.~T. Shen, and L.~Van~Gool,
  ``Generative domain-migration hashing for sketch-to-image retrieval,'' in
  \emph{ECCV}, 2018.

\bibitem{verma2019generative}
V.~K. Verma, A.~Mishra, A.~Mishra, and P.~Rai, ``Generative model for zero-shot
  sketch-based image retrieval,'' in \emph{CVPRW}, 2019.

\bibitem{li2019zero}
J.~Li, Z.~Ling, L.~Niu, and L.~Zhang, ``Zero-shot sketch-based image retrieval
  with structure-aware asymmetric disentanglement,'' \emph{arXiv preprint
  arXiv:1911.13251}, 2019.

\bibitem{tursun2021efficient}
O.~Tursun, S.~Denman, S.~Sridharan, E.~Goan, and C.~Fookes, ``An efficient
  framework for zero-shot sketch-based image retrieval,'' \emph{arXiv preprint
  arXiv:2102.04016}, 2021.

\bibitem{deng2020progressive}
C.~Deng, X.~Xu, H.~Wang, M.~Yang, and D.~Tao, ``Progressive cross-modal
  semantic network for zero-shot sketch-based image retrieval,'' \emph{IEEE
  Transactions on Image Processing}, vol.~29, pp. 8892--8902, 2020.

\bibitem{zhu2020ocean}
J.~Zhu, X.~Xu, F.~Shen, R.~K.-W. Lee, Z.~Wang, and H.~T. Shen, ``Ocean: A dual
  learning approach for generalized zero-shot sketch-based image retrieval,''
  in \emph{2020 IEEE International Conference on Multimedia and Expo
  (ICME)}.\hskip 1em plus 0.5em minus 0.4em\relax IEEE, 2020, pp. 1--6.

\bibitem{simonyan2014very}
K.~Simonyan and A.~Zisserman, ``Very deep convolutional networks for
  large-scale image recognition,'' \emph{arXiv}, 2014.

\bibitem{chaudhuri2021interband}
U.~Chaudhuri, S.~Dey, M.~Datcu, B.~Banerjee, and A.~Bhattacharya, ``Interband
  retrieval and classification using the multilabeled sentinel-2 bigearthnet
  archive,'' \emph{IEEE Journal of Selected Topics in Applied Earth
  Observations and Remote Sensing}, vol.~14, pp. 9884--9898, 2021.

\bibitem{chen2017beyond}
W.~Chen, X.~Chen, J.~Zhang, and K.~Huang, ``Beyond triplet loss: a deep
  quadruplet network for person re-identification,'' in \emph{CVPR}, 2017, pp.
  403--412.

\end{thebibliography}

\end{document}